\documentclass[letterpaper]{article} % DO NOT CHANGE THIS
\usepackage{aaai25}  % DO NOT CHANGE THIS
\usepackage{times}  % DO NOT CHANGE THIS
\usepackage{helvet}  % DO NOT CHANGE THIS
\usepackage{courier}  % DO NOT CHANGE THIS
\usepackage[hyphens]{url}  % DO NOT CHANGE THIS
\usepackage{graphicx} % DO NOT CHANGE THIS
\def\UrlFont{\rm}  % DO NOT CHANGE THIS
\usepackage{natbib}  % DO NOT CHANGE THIS AND DO NOT ADD ANY OPTIONS TO IT
\usepackage{caption} % DO NOT CHANGE THIS AND DO NOT ADD ANY OPTIONS TO IT
\usepackage{algorithm}
\usepackage{algorithmic}

\usepackage{newfloat}
\usepackage{listings}
\DeclareCaptionStyle{ruled}{labelfont=normalfont,labelsep=colon,strut=off} % DO NOT CHANGE THIS
\floatstyle{ruled}
\newfloat{listing}{tb}{lst}{}
\floatname{listing}{Listing}
\usepackage{booktabs,pifont,pythonhighlight}
\newcommand{\cmark}{\ding{51}}%
\newcommand{\xmark}{\ding{55}}%

\title{RLLTE: Long-Term Evolution Project of Reinforcement Learning}
\author{
    	Mingqi Yuan\textsuperscript{1}, Zequn Zhang\textsuperscript{2,5}, Yang Xu\textsuperscript{3}, Shihao Luo\textsuperscript{4}, Bo Li\textsuperscript{1}, Xin Jin\textsuperscript{2\thanks{Corresponding author}}, Wenjun Zeng\textsuperscript{2} \\
}
\affiliations{
    \textsuperscript{1}Department of Computing, The Hong Kong Polytechnic University, China\\
	\textsuperscript{2}Ningbo Institute of Digital Twin, Eastern Institute of Technology, Ningbo, China \\
	\textsuperscript{3}School of Industrial Engineering, Purdue University, USA \\
	\textsuperscript{4}Shenzhen Dajiang Innovation Technology Co., Ltd, China\\ 
    \textsuperscript{5}EEIS, University of Science and Technology of China, China \\
	mingqi.yuan@connect.polyu.hk, xu1720@purdue.edu
}

\usepackage{bibentry}
\begin{document}

\maketitle

\begin{abstract}
We present RLLTE: a long-term evolution, extremely modular, and open-source framework for reinforcement learning (RL) research and application. Beyond delivering top-notch algorithm implementations, RLLTE also serves as a toolkit for developing algorithms. More specifically, RLLTE decouples the RL algorithms completely from the exploitation-exploration perspective, providing a large number of components to accelerate algorithm development and evolution. In particular, RLLTE is the first RL framework to build a comprehensive ecosystem, which includes model training, evaluation, deployment, benchmark hub, and large language model (LLM)-empowered copilot. RLLTE is expected to set standards for RL engineering practice and be highly stimulative for industry and academia. Our documentation, examples, and source code are available at \url{https://github.com/RLE-Foundation/rllte}. 

\end{abstract}

    % \link{Code}{https://aaai.org/example/code}
    % \link{Datasets}{https://aaai.org/example/datasets}
    % \link{Extended version}{https://aaai.org/example/extended-version}

\section{Introduction}
Reinforcement learning (RL) has emerged as a highly significant research topic due to its remarkable achievements in diverse fields \cite{mnih2015human, silver2017mastering, mankowitz2023faster}. However, the efficient and reliable engineering implementation of RL algorithms remains a long-standing challenge. These algorithms often possess sophisticated structures, where minor code variations can substantially influence their practical performance. To tackle this problem, several open-source projects were proposed to offer reference implementations of popular RL algorithms, such as stable-baselines3 (SB3) \cite{raffin2021stable}, Tianshou \cite{weng2022tianshou}, and CleanRL \cite{huang2022cleanrl}. Despite their achievements, most of the existing benchmarks simply concentrate on providing algorithm implementations rather than constructing a comprehensive ecosystem (e.g., model evaluation, deployment, and exhaustive benchmark testing data). Moreover, they also set a high threshold for new developers, which hinders community collaboration.

Inspired by the discussions above, we propose \textbf{RLLTE}, a long-term evolution, extremely modular, and open-source framework of RL. RLLTE is expected to produce a positive impact on both academic research and industry applications.

\section{Architecture}

\begin{figure}[h!]
	\centering
	\includegraphics[width=\linewidth]{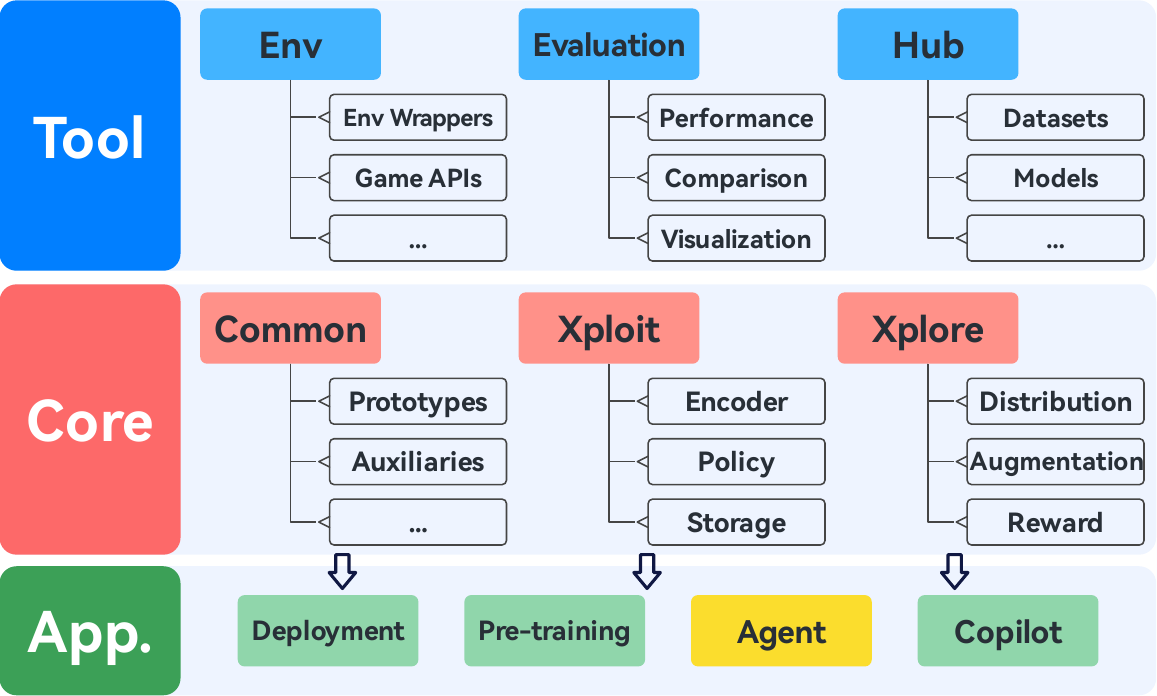}
	\caption{Overview of the architecture of RLLTE.}
	\label{fig:arch}
\end{figure}

\begin{table*}[h!]
\centering
\fontsize{9pt}{10pt}\selectfont
\tabcolsep=2.9pt
\begin{tabular}{c|cccccccccccc}
\toprule
\textbf{Framework} & \textbf{\begin{tabular}[c]{@{}c@{}}Number\\ of Algo.\end{tabular}} & \textbf{Modularized} & \textbf{Decoupling} & \textbf{Backend} & \textbf{\begin{tabular}[c]{@{}c@{}}Custom\\ Env.\end{tabular}} & \textbf{\begin{tabular}[c]{@{}c@{}}Custom\\ Module\end{tabular}} & \textbf{\begin{tabular}[c]{@{}c@{}}Data\\ Aug.\end{tabular}} & \textbf{\begin{tabular}[c]{@{}c@{}}Data\\ Hub\end{tabular}} & \textbf{Deploy.} & \textbf{Eval.} & \textbf{\begin{tabular}[c]{@{}c@{}}Multi-\\ Device\end{tabular}} & \textbf{Doc.} \\ \hline
Baselines          & 9                                                                  & \xmark               & -                   & TF               & \cmark(gym)                                                    & -                                                                & \xmark                                                       & -                                                           & \xmark           & \xmark         & \xmark                                                           & \xmark        \\
SB3                & 7                                                                  & \xmark               & -                   & PyTorch          & \cmark(gymnasium)                                              & -                                                                & -                                                            & \cmark                                                      & \xmark           & \xmark         & \xmark                                                           & \cmark        \\
CleanRL            & 9                                                                  & \xmark               & \xmark              & PyTorch          & \xmark                                                         & \cmark                                                           & -                                                            & \cmark                                                      & \xmark           & \xmark         & \xmark                                                           & \cmark        \\
Ray/rllib          & 16                                                                 & \cmark               & -                   & TF/PyTorch       & \cmark(gym)                                                    & -                                                                & -                                                            & -                                                           & \xmark           & \xmark         & \xmark                                                           & \cmark        \\
rlpyt              & 11                                                                 & \cmark               & \xmark              & PyTorch          & \xmark                                                         & -                                                                & \xmark                                                       & -                                                           & \xmark           & \xmark         & \xmark                                                           & \cmark        \\
Tianshou           & 20                                                                 & \cmark               & -                   & PyTorch          & \cmark(gymnasium)                                              & \xmark                                                           & -                                                            & -                                                           & \xmark           & \xmark         & \xmark                                                           & \cmark        \\
ElegantRL          & 9                                                                  & \cmark               & -                   & PyTorch          & \cmark(gym)                                                    & \xmark                                                           & \xmark                                                       & -                                                           & \xmark           & \xmark         & \xmark                                                           & \cmark        \\
SpinningUp         & 6                                                                  & \xmark               & \xmark              & PyTorch          & \cmark(gym)                                                    & \xmark                                                           & \xmark                                                       & -                                                           & \xmark           & \xmark         & \xmark                                                           & \cmark        \\
ACME               & 14                                                                 & \cmark               & \xmark              & TF/JAX           & \cmark(dm\_env)                                                & \xmark                                                           & \xmark                                                       & -                                                           & \xmark           & \xmark         & \xmark                                                           & \cmark        \\
Torch/rl           & 15                                                                 & \cmark               & \cmark              & PyTorch          & \cmark(gymnasium)                                              & \cmark                                                           & \xmark                                                       & \xmark                                                      & \xmark           & \xmark         & \cmark                                                           & \cmark        \\
RLLTE              & 13$\nearrow$                                                       & \cmark               & \cmark              & PyTorch          & \cmark(gymnasium)                                              & \cmark                                                           & \cmark                                                       & \cmark                                                      & \cmark           & \cmark         & \cmark                                                           & \cmark        \\ \bottomrule
\end{tabular}
\vskip -0.05in
\caption{Comparison with existing RL projects. \textbf{Modularized}: The project adopts a modular design with reusable components. \textbf{Decoupling}: The project supports algorithm decoupling and module replacement. \textbf{Custom Env.}: Support custom environments? Since Gym \citep{brockman2016openai} is no longer maintained, it is critical to make the project adapt to Gymnasium \citep{towers2024gymnasium}, which is a maintained fork of Gym. \textbf{Custom Module}: Support custom modules? \textbf{Data Aug.}: Support data augmentation techniques like intrinsic reward shaping and observation augmentation? \textbf{Data Hub}: Have a data hub to store benchmark data? \textbf{Deploy.}: Support model deployment? \textbf{Eval.}: Support model evaluation? \textbf{Multi-Device}: Support hardware acceleration of different computing devices (e.g., GPU and NPU)? Note that the short line represents partial support.}
\label{tb:comp func}
\vskip -0.25in
\end{table*}

\begin{figure}[h!]
	\begin{center}
		\includegraphics[width=\linewidth]{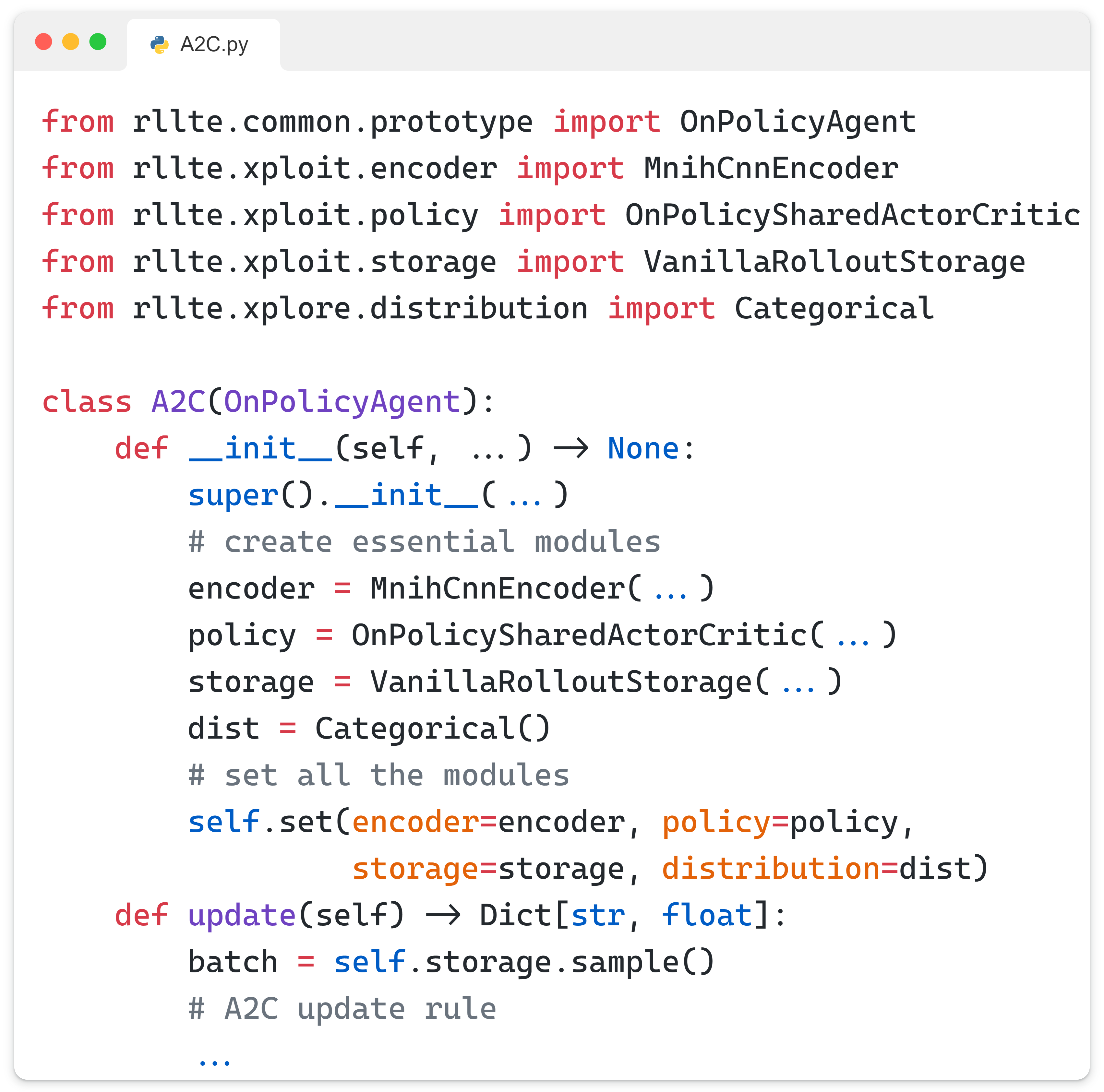}
	\end{center}
    \vskip -0.15in
	\caption{Implement the A2C algorithm with dozens of lines of code, and the complete code example can be found in the official code repository.}
	\label{fig:fast}
    \vskip -0.3in
\end{figure}

Figure~\ref{fig:arch} illustrates the overall architecture of RLLTE, which consists of the core layer, application layer, and tool layer. We summarize the highlighted features of RLLTE as follows:

\textbf{Module-oriented.} RLLTE decouples RL algorithms from the \emph{exploitation-exploration} perspective and breaks them down into minimal primitives, such as \emph{encoder} for feature extraction and \emph{storage} for archiving and sampling experiences. RLLTE offers a rich selection of modules for each primitive, enabling developers to utilize them as building blocks for constructing algorithms (see Figure~\ref{fig:fast}). As a result, the focus of RLLTE shifts from specific algorithms to providing more handy modules like PyTorch. In particular, each module in RLLTE is customizable and plug-and-play, empowering users to develop their own modules. This decoupling process also contributes to advancements in interpretability research, allowing for a more in-depth exploration of RL algorithms.
	
\textbf{Long-term evolution.} RLLTE is a long-term evolution project, continually involving advanced algorithms and tools in RL. RLLTE will be updated based on the following tenet: (i) generality; (ii) improvements in generalization ability and sample efficiency; (iii) excellent performance on recognized benchmarks; (iv) promising tools for RL. Therefore, this project can uphold the right volume and high-quality resources, thereby inspiring more subsequent projects.

\textbf{Data augmentation.} Recent approaches have introduced data augmentation techniques at the \emph{observation} and \emph{reward} levels to improve the sample efficiency and generalization ability of RL agents, which are cost-effective and highly efficient \cite{pathak2017curiosity, raileanu2020ride,laskin2020curl,laskin2020reinforcement}. In line with this trend, RLLTE incorporates built-in support for data augmentation operations and offers a wide range of observation augmentation modules and intrinsic reward modules.
	
\textbf{Abundant ecosystem.} RLLTE considers the needs of both academia and industry and develops an abundant project ecosystem. For instance, RLLTE designed an evaluation toolkit to provide statistical and reliable metrics for assessing RL algorithms. Additionally, the deployment toolkit enables the seamless execution of models on various inference devices. In particular, RLLTE attempts to introduce the large language model (LLM) to build an intelligent copilot for RL research and applications.
	
\textbf{Comprehensive benchmark data.} Existing RL projects typically conduct testing on a limited number of benchmarks and often lack comprehensive training data, including learning curves and test scores. While this limitation is understandable, given the resource-intensive nature of RL training, it hampers the advancement of subsequent research. To address this issue, RLLTE has established a data hub utilizing the Hugging Face platform. This data hub provides extensive testing data for the included algorithms on widely recognized benchmarks. By offering complete and accessible testing data, RLLTE will facilitate and accelerate future research endeavors in RL.
	
\textbf{Multi-hardware support.} RLLTE has been thoughtfully designed to accommodate diverse computing hardware configurations, including graphic processing units (GPUs) and neural network processing units (NPUs), in response to the escalating global demand for computing power. This flexibility enables RLLTE to support various computing resources, ensuring optimal trade-off of performance and scalability for RL applications.

Finally, we provide a systematic comparison of architecture and functionality between RLLTE and other representative RL projects in Table~\ref{tb:comp func}. 

\vfill

\section{Conclusion}
This paper introduces RLLTE, a novel, long-term evolution, extremely modular, and open-source framework for advancing RL research and applications. RLLTE provides a comprehensive ecosystem that enables developers to perform task design, model training, evaluation, and deployment seamlessly within one framework. This framework is expected to be highly stimulative for both academia and industry.

\clearpage
\newpage

\section*{Acknowledgments}
This work is supported by the HKSAR Research Grants Council under Grant No. PolyU 15224823, the Guangdong Basic and Applied Basic Research Foundation under Grant No. 2023A1515010592, the NSFC under Grant No. 62302246, and ZJNSFC under Grant No. LQ23F010008. We thank the HPC center at the Eastern Institute of Technology, Ningbo, and the Ningbo Institute of Digital Twin for providing their GPU computing platform for testing.

\bibliography{aaai25}

\end{document}